\documentclass{article}

\usepackage[final, nonatbib]{neurips_2020}

\usepackage[utf8]{inputenc} %
\usepackage[T1]{fontenc}    %
\usepackage{hyperref}       %
\usepackage{url}            %
\usepackage{booktabs}       %
\usepackage{amsfonts}       %
\usepackage{nicefrac}       %
\usepackage{microtype}      %

\usepackage[ruled,vlined]{algorithm2e}
\usepackage[numbers]{natbib}
\usepackage{graphicx}
\usepackage{caption}
\usepackage{subcaption}
\graphicspath{{./diagrams/}}
\usepackage{amsmath}

\usepackage{lscape}
\usepackage{rotating}

\usepackage[draft]{todonotes}

\newcommand{\fV}{\mathbf{f}}
\newcommand{\aV}{\mathbf{a}}
\newcommand{\aM}{\mathbf{A}}
\newcommand{\xV}{\mathbf{x}}
\newcommand{\xM}{\mathbf{X}}

\newcommand{\K}{\mathbf{K}}

\newcommand{\dif}{\mathop{}\!\mathrm{d}}

\newcommand{\gaussianDistTwo}[2]{\mathcal{N}\left(#1, #2\right)}
\newcommand{\expectedDist}[2]{\left\langle#1\right\rangle_{#2}}
\newcommand{\model}{\mathbf{M}}
\newcommand{\data}{\mathcal{D}}
\newcommand{\mVbar}{\mathbf{\bar{m}}}
\newcommand{\pM}{\mathbf{P}}

\newcommand{\zM}{\mathbf{Z}}
\newcommand{\wM}{\mathbf{W}}
\newcommand{\mV}{\mathbf{m}}
\newcommand{\uV}{\mathbf{u}}
\newcommand{\sM}{\mathbf{S}}
\newcommand{\diag}[1]{\text{diag}\left(#1\right)}

\newcommand{\ie}{\textit{i.e.}}
\newcommand{\eg}{\textit{e.g.}}
\newcommand{\us}{AOE}
\newcommand{\problem}{MSPS}

\title{Model Selection for Production System via Automated Online Experiments}

\author{%
  Zhenwen Dai \\
  Spotify\\
  \texttt{zhenwend@spotify.com} \\
  \And
  Praveen Chandar \\
  Spotify \\
  \texttt{praveenr@spotify.com} \\
  \AND
  Ghazal Fazelnia \\
  Spotify \\
  \texttt{ghazalf@spotify.com} \\
  \And
  Ben Carterette \\
  Spotify \\
  \texttt{benjaminc@spotify.com} \\
  \And
  Mounia Lalmas-Roelleke \\
  Spotify \\
  \texttt{mounial@spotify.com} \\
}

\begin{document}

\maketitle

\begin{abstract}

A challenge that machine learning practitioners in the industry face is the task of selecting the best model to deploy in production. As a model is often an intermediate component of a production system, online controlled experiments such as A/B tests yield the most reliable estimation of the effectiveness of the whole system, but can only compare two or a few models due to budget constraints. 
We propose an automated online experimentation mechanism that can efficiently perform model selection from a large pool of models with a small number of online experiments.
We derive the probability distribution of the metric of interest that contains the model uncertainty from our Bayesian surrogate model trained using historical logs.
Our method efficiently identifies the best model by sequentially selecting and deploying a list of models from the candidate set that balance exploration-exploitation.
Using simulations based on real data, we demonstrate the effectiveness of our method on two different tasks.
\end{abstract}

\section{Introduction}

Evaluating the effect of individual changes to machine learning (ML) systems such as choice of algorithms, features, \textit{etc.}, is the key to growth in many internet services and industrial applications. 
Practitioners are faced with the decision of choosing one model from several candidates to deploy in production. This can be viewed as a model selection problem. Classical model selection paradigms such as cross-validation consider ML models in isolation and focus on selecting the model with the best predictive power on unseen data. This approach does not work well for modern industrial ML systems, as such a system usually consists of many individual components and a ML model is only one of them. The metric of interest often depends on uncontrollable factors such as users' responses. Only optimizing the predictive power of the ML model would not lead to a better metric of the overall system.
Instead, randomized experiments (also known as "A/B tests") are considered as the gold-standard for evaluating system changes~\cite{Kohavi2007} as they provide a more direct measure of the metric. However, only a few variants of the ML model can be tested using randomized experiments as they are time-consuming to conduct and have resource constraints (such as the number of active users, \textit{etc.}). Furthermore, deploying bad systems can lead to catastrophic consequences. 

An alternative approach is to exploit log data collected under the production system to estimate the metric of interest if a different ML model is deployed. Typical methods include developing offline measures or simulators that model users' behavior~\cite{Hofmann2016,Sanderson2010}, and replaying the recorded decisions with a probability ratio correction between the new and previous model, which is referred to as off-policy evaluation (OPE)~\citep{PrecupEtAl2000, DudikEtAl2014, MehrdadEtAl2018, YaoEtAl2018, VlassisEtAl2019, AlexanderEtAl2019}. A big challenge faced by these methods is the selection bias of the log data. As a consequence, these methods work well when the considered model behaves similar to the logging model, but the effectiveness deteriorates quickly when the considered model behaves increasingly differently from the logging model.

To overcome the selection bias, we suggest to include the data collection process into the model selection approach. We propose a new framework of model selection for production system, where the best model is selected via deploying  a sequence of models online. This allows deploying the model that can provide maximum information, iteratively refining the belief about the candidate models and efficiently identifying the model that leads to the best metric for the overall system.
Concretely, we target at a specific but widely existing scenario, in which the metric of interest can be decomposed into an average of immediate feedback, \eg{}, the click-through rate in recommender systems. We develop a Bayesian surrogate model that can efficiently digest the collected data from online experiments and derive the probability distribution of the metric of interest based on the surrogate model. The model to deploy is selected by balancing the exploration-exploitation trade-off. 
Comparing with A/B testing, our approach can perform model selection from a large pool of candidates by using not only the recorded metric but also the log data about individual user interactions. Comparing with OPE, our approach provides  more accurate estimation of model performance by avoiding the selection bias through controlling the data collection process. Overall, our approach correctly identifies the best model even if it behaves very differently from the one in production.

\section{Model selection with automated online experiments}\label{sec:problem_statement}

We define the problem of model selection for production system (\problem{}) as follows: given a set of candidate models $\model_i \in \mathcal{M}$ and an online budget, the goal is to identify model $\model^*$ with maximum utility of the overall system: 
\begin{equation}
\model^* = \arg \max_{\model_i \in \mathcal{M}} v(\model_i).
\end{equation}
In this work, we focus on the scenario where a model takes an input representation $\xV$, returns a decision $\aV$ while observing an immediate feedback for each individual decision. 
The utility associated with a given model $\model_i$ is influenced by immediate feedback, which could be an indirect and complex relationship such as the relation between profit margin and user clicks.
Here, we restrict our focus to the cases where the utility has an additive relation with immediate feedback and refer to it as \emph{accumulative metric}. 
The above setting is common in the industry; for example, in recommender systems, the inputs $\xV$ are users or context (user representation, time of the request, \textit{etc.}), the decisions are the choice of recommendation, and the accumulative metric could be a metric such as total consumption, which is the sum of consumption associated with individual recommendations. 

A model can be represented as a probability distribution of the decision conditioned on the input $p(\aV|\xV)$, where a deterministic system simply results in a delta distribution. The distribution of the inputs to the model represented as $p(\xV)$ is typically unknown. The accumulative metric for a given model $\model_i$ can be defined as, 
\begin{equation} 
v(\model_i) = \int_{\mathcal{X}} \int_{\mathcal{A}} m\, p(m | \aV, \xV) p(\aV|\xV, \model_i) p(\xV) \dif \aV \dif \xV, \label{eqn:def_expected_reward} 
\end{equation} 
where the integration is over the space of input $\xV \in \mathcal{X}$ and the space of decision $\aV \in \mathcal{A}$. The accumulative metric is defined as an expectation of immediate feedback with respect to the distribution of input and decisions conditioned on individual inputs. Unfortunately, the accumulative metric is not tractable, because both the distribution of input $p(\xV)$ and the distribution of immediate feedback $p(m | \aV, \xV)$ are unknown. 

With a production system, the information about the accumulative metric can be collected by deploying the model of interest in production and let it react to real traffic and record the corresponding accumulative metric.
The collected data from such a deploy consist of a recorded accumulative metric, in our case $\hat{v} = \frac{1}{N}\sum_i m_i$, and a set of interactions $\mathcal{D} = \{(m_i, \aV_i, \xV_i)\}_{i=1}^{N}$.  
In this work, we define \problem{} as a sequential optimal decision problem. A \problem{} method iteratively chooses a model from the candidate set to deploy online for data collection with the aim of identifying the model with the best accumulative metric in the fewest number of online experiments. A model deployment is a expensive process, as each deployment takes a long time, and only a small number of models can be deployed in parallel due to the limited number of users and the affordable degradation in service quality. Global optimization methods like Bayesian optimization (BO)~\citep{SnoekEtAl2012} do not work well in this setting, because BO requires the search space to be represented  in a relatively low dimensional space but embedding the model candidate sets (especially models of different types) into a low-dimensional space is non-trivial. Unlike BO methods that only take into account the accumulative metric from online experiments, our approach takes advantage of the full log data by training a Bayesian surrogate model. The uncertainty of the surrogate model is then used to balance between exploring an uncertain but potentially good choice and exploiting a known one.

\section{Bayesian surrogate for accumulative metric}\label{sec:surrogatemodel}

Instead of using the recorded accumulative metric from online experiments, we propose to estimate it from its definition in (\ref{eqn:def_expected_reward}). In this formulation, $p(\aV|\xV, \model_i)$ is known and $p(\xV)$ could be replaced with an empirical distribution, therefore, the key is to capture the distribution of the immediate feedback $p(m|\aV, \xV)$. The data collected from online experiments contains lots of data points about this distribution. This allows us to build a Bayesian surrogate model for the immediate feedback.

\subsection{Gaussian process surrogate model}\label{sec:gp_surrogate_model}

We propose to use a Gaussian process (GP) as the surrogate model for the distribution of the immediate feedback. There is often stochasticity in the immediate feedback data including the intrinsic stochasticity in human interactions, e.g., some random reactions from a user, as well as the information that is not available to the production system. To accommodate this stochasticity, we divide the Bayesian surrogate model into two parts: (i) a GP model that captures the ``noise-free" component of the immediate feedback, denoted as $p(f|\aV,\xV)$; (ii) a noise distribution used to absorb all the stochasticity that cannot be explained by $\xV$ and $\aV$, denoted as $p(m|f)$.
When the immediate feedback is a continuous value, we use a Gaussian noise distribution. The resulting surrogate model can be written as,
\begin{equation}
m = f(\aV, \xV)+\epsilon , \quad f \sim \mathcal{GP}(0, k(\cdot, \cdot)), \quad \epsilon \sim \mathcal{N}(0, \sigma^2),
\end{equation}
where the GP has zero mean and a covariance function $k(\cdot, \cdot)$. Stationary covariance functions are the most common choices, such as the radial basis function (RBF) and the Mat\'{e}rn covariance functions. Note that the distribution of the immediate feedback $p(m|\aV, \xV)$ is independent of the choice of candidate models. This allows us to train a single surrogate model and use it to score all the candidate models.

In some use cases, the inputs $\xV$ and/or the decisions $\aV$ may be categorical values, \eg, in recommender systems, the input may be a user ID and the decision may be an item ID, both of which are categorical values. The standard one-hot encoding is not a good representation for GP. Instead, we embed each unique ID as a latent variable in a low dimensional space, \eg{}, $\aV_k \in \mathbb{R}^Q, \aV_k \sim \mathcal{N}(0, \mathbf{I})$. This approach is closely related to variational multi-output GPs \citep{DaiEtAl2017}. Deep GPs \citep{DamianouLawrence2013, DaiEtAl2016} can be considered if the distribution of immediate feedback is heavily non-stationary.

With a surrogate model for the immediate feedback, we have all the pieces to estimate the accumulative metric from (\ref{eqn:def_expected_reward}). The integral is generally intractable but could be approximated by the methods like Monte Carlo sampling. Note that the resulting quantity $v(\model_i)$ is deterministic as all the involved probability distributions are integrated out. It can serve as an estimator for the accumulative metric but is \textit{unable} to be used for exploration-exploitation tradeoff. In order to construct an efficient \problem{} method, we need to represent the accumulative metric as a random variable, of which the uncertainty reflects the current belief of its value according to the surrogate model, which is often referred to as \textit{model uncertainty}.

\subsection{Estimation of the accumulative metric}

To derive the accumulative metric as a random variable that reflects model uncertainty, we first need to remove the uncertainty from the noise distribution, which corresponds to \textit{aleatoric uncertainty}. This is particularly crucial for the case of a binary immediate feedback, which will be explained in the next section. 
Firstly, we derive the expected immediate feedback from the noise distribution, i.e., $\bar{m} = \expectedDist{m}{p(m|f)}$.
In the case of a normal noise distribution, the expected immediate feedback is the mean of the noise distribution,
$\bar{m} = \int m \mathcal{N}(m; f, \sigma^2) \dif m = f$. Then, we derive the predictive expected immediate feedback from a inferred GP surrogate model by a change of random variable, 
$p(\mVbar | \aM, \xM, \mathcal{D}) = p(\fV|_{\fV=\mVbar} | \aM, \xM, \data)$, 
where $p(\fV | \aM, \xM, \data)$ is the noise-free predictive distribution from GP conditioned on the collected data via online experiments.

Consider a list of inputs $\xM=(\xV_1, \ldots, \xV_T)$ and the decision space $\mathcal{A}$ being discrete, denoted as $\aM = (\aV_1, \ldots, \aV_K)$. Given a model $\model_i$, the distribution of the model can be represented as a matrix $\pM \in [0, 1]^{K \times T}$, where each entry $p_{ij} = p(\aV_i | \xV_j)$.
The accumulative metric is defined as the sum of immediate feedback weighted by the inputs and decision probabilities. This allows us to derive the accumulative metric as a random variable $\hat{v} | \model_i, \mathcal{D}$,
\begin{equation}
\hat{v} | \model_i, \mathcal{D} = \frac{1}{T} \pM_:^\top \mVbar, \quad \mVbar \sim p(\mVbar | \aM, \xM, \mathcal{D}),
\label{eqn:gp_reg_accum_metric}
\end{equation}
where  the subscript $:$ denotes the vectorization of a matrix and $\mVbar$ is the vector of expected immediate feedback corresponding to the combinatorial of $\xM$ and $\aM$, denoted as $\wM = ((\aV_1, \xV_1), \ldots, (\aV_K, \xV_1), \ldots, (\aV_K, \xV_T))$. 
As the change of random variable in (\ref{eqn:gp_reg_accum_metric}) is a linear operation, the resulting random variable $\hat{v}$ is jointly GP distributed with $\mVbar$. It turns out that the resulting distribution $p(\hat{v}|\model_i, \mathcal{D})$ can be derived in closed-form,
\begin{equation}
p(\hat{v}|\model_i, \mathcal{D}) = \gaussianDistTwo{\frac{1}{T}\pM_: ^\top  \K_* \K^{-1} \mV}{\frac{1}{T} \pM_:^\top ( \K_{**} - \K_* \K^{-1} \K_*^\top)\pM_:},
\end{equation}
where $\mV$ is the recorded immediate feedback in $\data$, $\K$ is the covariance matrix among the observed data $\data$, $\K_*$ is the cross-covariance matrix between $\wM$ and $\data$ and $\K_{**}$ is the covariance among $\wM$.

Note that the expectation of the random variable $\hat{v}$ recovers the accumulative metric estimator in (\ref{eqn:def_expected_reward}), \ie, $v(\model_i) = \int \hat{v} p(\hat{v} | \model_i, \data) \dif \hat{v}$. As the probability distributions of inputs and decisions are represented in the matrix $\pM$ and the uncertainty from the noise distribution is removed, the uncertainty in $\hat{v}$ is a result of the model uncertainty of the GP surrogate model, which is crucial for the exploration-exploitation tradeoff.

For a real world problem, $\data$ often contains lots of data points, for which the cubic complexity of exact GP inference is too expensive. For scalability, we use the variational sparse GP approximation \citep{Titsias2009}. It augments the original data with a set of pseudo data $\uV$ at the corresponding locations $\zM$. Such an augmentation does not change the original model distribution $p(\fV|\aM, \xM) = \int p(\fV|\uV, \aM, \xM, \zM) p(\uV|\zM) \dif \uV$. With an efficient variational lower bound, the computational complexity reduces from $O(N^3)$ to $O(NC^2)$, where $C$ is the number of pseudo data. The inference result of sparse GP is often represented by the variational posterior of the pseudo data, denoted as $q(\uV) = \mathcal{N}(\mV_{\uV}, \sM_{\uV})$.
With sparse GP approximation, the distribution $p(\hat{v}|\model_i, \mathcal{D})$ becomes
\begin{equation}
p(\hat{v}|\model_i, \mathcal{D}) = \mathcal{N}\left( \frac{1}{T}\pM_: ^\top  \K_{*\uV} \K_{\uV\uV}^{-1} \mV_{\uV}, \frac{1}{T} \pM_:^\top ( \K_{**} - \K_{*\uV} (\K_{\uV\uV}^{-1} - \K_{\uV\uV}^{-1}  \sM_{\uV} \K_{\uV\uV}^{-1} )\K_{*\uV}^\top)\pM_: \right),
\label{eqn:sparse_gp_reg_accum_metric}
\end{equation}
where $\K_{\uV\uV}$ is the covariance matrix among the pseudo data and $\K_{*\uV}$ is the cross-covariance matrix between $\wM$ and the pseudo data.

For a large problem, the variance calculation in (\ref{eqn:sparse_gp_reg_accum_metric}) can also be very expensive as  $\K_{**}$ is a $KT$-by-$KT$ matrix. For efficient computation, we use a FITC approximation \citep{SnelsonGhahramani2006} at prediction time, \ie, $p_{\text{FITC}}(\fV|\uV, \aM, \xM, \zM) = \mathcal{N}(\K_{\fV\uV}\K_{\uV\uV}^{-1}\uV, \mathbf{\Lambda} )$, where $\mathbf{\Lambda} =\diag{\K_{\fV\fV} -\K_{\fV\uV}\K_{\uV\uV}^{-1} \K_{\fV\uV}^\top}$ and  $\diag{\cdot}$ makes a matrix into a diagonal matrix by letting off-diagonal entries be zero. Note that, although the conditional distribution $p(\fV|\uV)$ is independent among the entries of $\fV$, the resulting distribution $p(\mVbar | \aM, \xM, \mathcal{D})$ is still correlated due to the correlation from the pseudo data.
With the FITC approximation, the mean $p(\hat{v}|\model_i, \mathcal{D})$ remains to be the same, while the variance becomes $\frac{1}{T} \pM_:^\top ( \mathbf{\Lambda} + \K_{*\uV}  \K_{\uV\uV}^{-1}  \sM_{\uV} \K_{\uV\uV}^{-1} \K_{*\uV}^\top)\pM_:$, in which only the diagonal entries of $\K_{**}$  needs to be computed.

\subsection{Binary immediate feedback}

In industrial use cases, binary immediate feedback is widely used because it is easy to calculate and easy to interpret by human, \eg{}, whether a user has responded to a shown item, whether a customer has purchased an item or whether a user has played a music or a movie. To apply our method to binary immediate feedback, we need to modify the GP surrogate model.

Firstly, we need to change the noise distribution to a Bernoulli distribution, $p(m|f) = \sigma(f)^m(1-\sigma(f))^{1-m}$,
where $\sigma(\cdot)$ is a link function that squashes the value of $f$ to be in $(0, 1)$. The most common link function is the logistic function. This makes the GP surrogate model become a GP binary classification model, of which the marginal likelihood is no longer closed-form. For both tractability and scalability, we use stochastic variational sparse GP approximation \citep{HensmanEtAl2013}, of which the intractable 1D integral in the variational lower bound is approximated by Gauss–Hermite quadrature.
Then, we derive the expected immediate feedback from the Bernoulli distribution, which is the probability of the immediate feedback being one, \ie, $\bar{m} = \sum_{m \in \{0, 1\}} m\,  p(m|f)  = \sigma(f)$.
The predictive expected immediate feedback from a inferred GP surrogate model can be derived by a change of random variable, 
\begin{equation}
p(\mVbar | \aM, \xM, \mathcal{D}) = p(\fV|_{\fV=\sigma^{-1}(\mVbar)}| \aM, \xM, \data) \begin{vmatrix}\frac{\dif \sigma^{-1}(\mVbar)}{\dif \mVbar} \end{vmatrix}.
\end{equation}
where $\sigma^{-1}(\cdot)$ is the inverse of the link function. Both $\sigma(\cdot)$ and $\sigma^{-1}(\cdot)$ are scalar functions. We use $\sigma^{-1}(\mVbar)$ to denote applying $\sigma^{-1}(\cdot)$ to the individual entries of $\mVbar$.
For binary immediate feedback, the random variable of the accumulative metric $\hat{v}$ defined in (\ref{eqn:gp_reg_accum_metric}) no longer has a closed form probability density function. 
Fortunately, we can efficiently sample from $p(\hat{v}|\model_i, \mathcal{D})$, by first drawing a sample from the ``noise-free" GP surrogate model and compute the sample of $\hat{v}$ according to (\ref{eqn:gp_reg_accum_metric}), \ie,
\begin{equation}
\hat{v}_i = \frac{1}{T} \pM_:^\top \sigma(\fV_i), \quad \fV_i \sim p(\fV | \aM, \xM, \mathcal{D}).
\end{equation}
Note that for binary immediate feedback, it is crucial to derive the random variable $\hat{v}$ from the expected immediate feedback $\bar{m}$ instead of the original immediate feedback $m$. Imagine that we derive $\hat{v}$ from $m$ by replacing $\mVbar$ with $\mV$. The consequence is that the variance of $\hat{v}$ will be at maximum if the expected immediate feedback $\mV$ equals to 0.5, no matter how small the model uncertainty is. In this case, the uncertain in $\hat{v}$ does not reflect the amount of unknowns in the surrogate model. Instead, deriving $\hat{v}$ from $\bar{m}$ can avoid this problem because the uncertainty from the noise distribution is excluded.

\section{Choosing the next online experiment} \label{sec:choosenext}

After deriving the probability distribution of the accumulative metric, we use an acquisition function to guide the choice of the next online experiment. We consider the acquisition functions $\alpha(\cdot)$ widely used in BO \eg~expected improvement (EI), probability of improvement (PI) and upper confidence bound (UCB). A major difference to BO is that the space of choices is no longer the same as the input space of the surrogate model. In our case, the space of choices are the set of candidate models, while the input to the surrogate model is the input to the ML model and the decision from the ML model. As a result, the acquisition functions designed by considering an extra hypothetical evaluation such as entropy search~\citep{HennigSchuler2012} cannot be used within our method. For binary feedback, the acquisition functions are not closed form. We use Monte Carlo sampling to compute the acquisition function by drawing samples from the distribution of the accumulative metric.

We name the resulting algorithm as automated online experimentation (\us{}), of which the overall procedure is shown in Algorithm~\ref{alg:algorithmflow}. We start with an initial dataset $\data_0$, which may be collected by deploying the model online. The model can be randomly chosen or chosen according to some domain knowledge or offline accuracy measure. Note that often the training data for the candidate models may be used to train the surrogate model as well. 
At each iteration, we first update surrogate model by inferring the variational posterior distribution as mentioned in Section~\ref{sec:surrogatemodel}. Then, we score all the candidate models with the acquisition function, which takes as inputs the distribution of the accumulative metric and select the model with the highest score. The selected model is deployed online for data collection. The collected data are augmented into the dataset for updating the surrogate model. We repeat this process until the online experiment budget is over. Then, the best model can be estimated from the latest surrogate model.

\begin{algorithm}[t]
\SetAlgoLined
\KwResult{Return the ML system with the highest accumulative metric}
Collect the initial data $\data_0$\;
 \While{Online experiment budget is not over}{
  Infer $p(\fV | \aM, \xM, \data_{t-1})$ with VI on surrogate model \;
  Identify $\model_t = \arg \max_{\model_i \in \mathcal{M}} \alpha(\model_i)$\;
  Deploy $\model_t$ and construct $\data_t$ by augmenting the collected data into $\data_{t-1}$ \;
 }
 \caption{model selection with automated online experiments (\us{})}\label{alg:algorithmflow}
\end{algorithm}

\section{Related Work}\label{sec:related_work}

Model selection~\citep{Bishop2006} is a classical topic in ML. The standard paradigm of model selection considers a model in insolation and aims at selecting a model that has the best predictive power for unseen data based on an offline dataset.
Common techniques such as cross-validation, bootstrapping, Akaike information criterion~\citep[AIC,][]{Akaike1974} and Bayesian information criterion~\citep[BIC,][]{Schwarz1978} have been widely used for scoring a model's predictive power based on a given dataset.
As scoring all the candidate models does not scale for complex problems, many recent works focus on tackling the problem of searching a large continuous and/or combinatorial space of model configurations, ranging from hyper-parameter optimization~\citep[HPO,][]{SnoekEtAl2012, AaronEtAl2019}, automatic statistician~\citep{LloydEtAl2014,GustavoEtAl2016, KimTeh2018, LuEtAl2018} to neural network architecture search~\citep[NAS,][]{ElskenEtAl2019}.  A more recent work~\citep{ChaiEtAl2019} jointly considers the scoring and searching problem for computational efficiency.
Online model selection~\citep{Sato2001, VidyaEtAl2019} is an extension of the standard model selection paradigm. It still treats a model in isolation but considers the online learning scenario, in which data arrive sequentially and the models are continuously updated. This is different to \problem{}, which views a model in the context of a bigger system and actively controls the data collection.

In reinforcement learning (RL), a model is considered as a decision mechanism, referred to as a \emph{policy}, and is evaluated for its associated accumulative rewards. Off-policy evaluation (OPE)~\citep{PrecupEtAl2000, DudikEtAl2014, MehrdadEtAl2018, YaoEtAl2018, VlassisEtAl2019, AlexanderEtAl2019} predicts the value of a new policy from an offline dataset logged by another policy. The definition of the value of a policy shares a similar format with the accumulative metric, which allows us to develop some baseline methods based on OPE. The major difference is that our method considers data collection as part of the selection process, which leads to the derivation of the probability distribution of the accumulative metric instead of an estimator of the value of a policy as in OPE.
Many works in RL also exploit the idea of Bayesian modeling and BO~\citep{GhavamzadehEngel2007, MohammadEtAl2016}, where GP is used to model the value function and obtain policy gradient via Bayesian quadrature. Lee et al.~\citep{LeeEtAl2019} use Bayesian models to represent the belief distribution over latent state space and perform off-policy update via a trust region method.
Recently, Letham and Bakshy~\citep{LethamBakshy2019} use multi-fidelity BO for policy search by correlating online and offline metrics. Their approach relies on the assumption that there exists an offline metric correlated with the online metric and  suffers from the limitation of BO, which requires the search space to be relatively low-dimensional.
Russo~\citep{Russo2016} considers the best-arm selection problem in contextual bandits, of which the used techniques are related to model selection but operate at a lower granularity.

Optimal experimental design is an area of research that focuses on techniques for efficient usage of limited resources in training models and data collections. Bayesian optimal experimental design tackles this problem by constructing a predictive model for possible experimental outcome, and seeks to optimize the expected information gain based on the posterior predictive estimation~\citep{chaloner1995bayesian,hernandez2014predictive,foster2019variational}. Rather than selecting model based on logging data, they take a parallel approach of optimizing data selection process based on the logging information, which has been successfully applied to various settings including bioinformatics~\citep{vanlier2012bayesian}, active learning~\citep{golovin2010near}, and neuroscience~\citep{shababo2013bayesian}.

\section{Experiment}

We demonstrate the performance of \us{} on automating online experiments for model selection. We construct two simulators based on real data to perform the evaluation since evaluation on a production system is not reproducible. 
We compare \us{} with five baseline methods: (i) directly applying BO on the collected accumulative metrics; (ii) using two OPE methods (importance sampling (IS) and doubly robust (DR)) to estimate the accumulative metrics at each iteration and greedily deploy the model with the best estimated metric, denoted as IS-g and DR-g respectively; and finally (iii) using two OPE methods to estimate the accumulative metrics with their empirical variance and choose the model to deploy according to an acquisition function (EI),  denoted as IS-EI and DR-EI, respectively. For the details of the baseline methods see the supplement material.

\begin{figure}[t]
     \centering
     \begin{subfigure}[b]{0.31\textwidth}
         \centering
         \includegraphics[width=\textwidth]{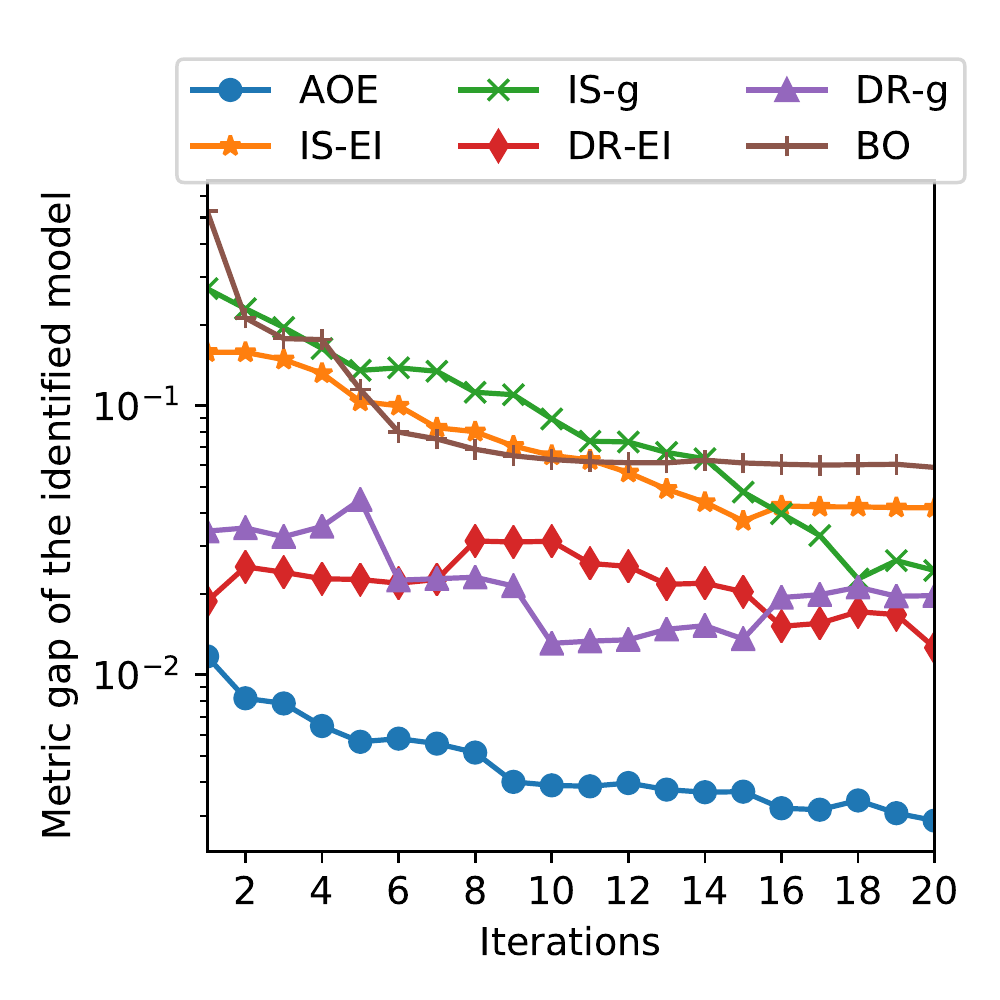}\vspace{-2.5mm}
         \caption{}
         \label{fig:svm_progress}
     \end{subfigure}
      \begin{subfigure}[b]{0.31\textwidth}
         \centering
         \includegraphics[width=\textwidth]{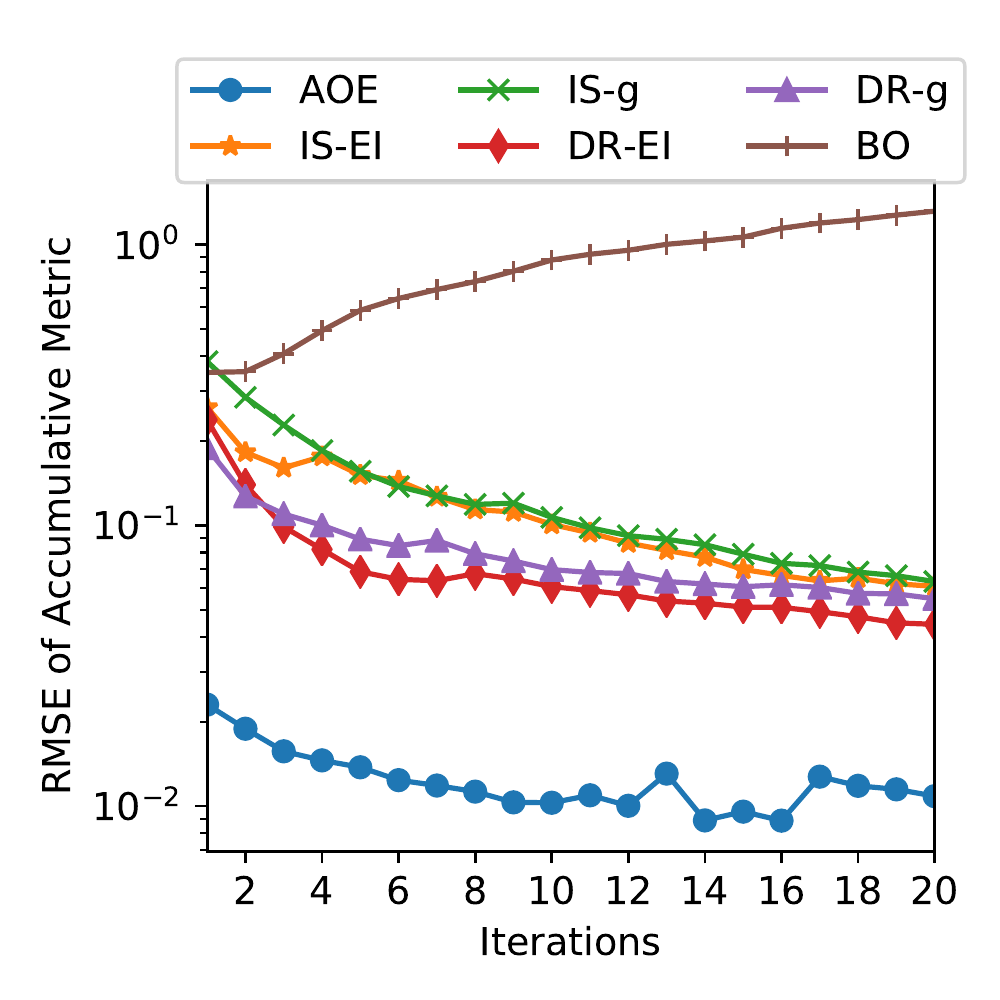}\vspace{-2.5mm}
         \caption{}
         \label{fig:svm_rmse_progress}
     \end{subfigure}
      \begin{subfigure}[b]{0.32\textwidth}
         \centering
         \includegraphics[width=\textwidth]{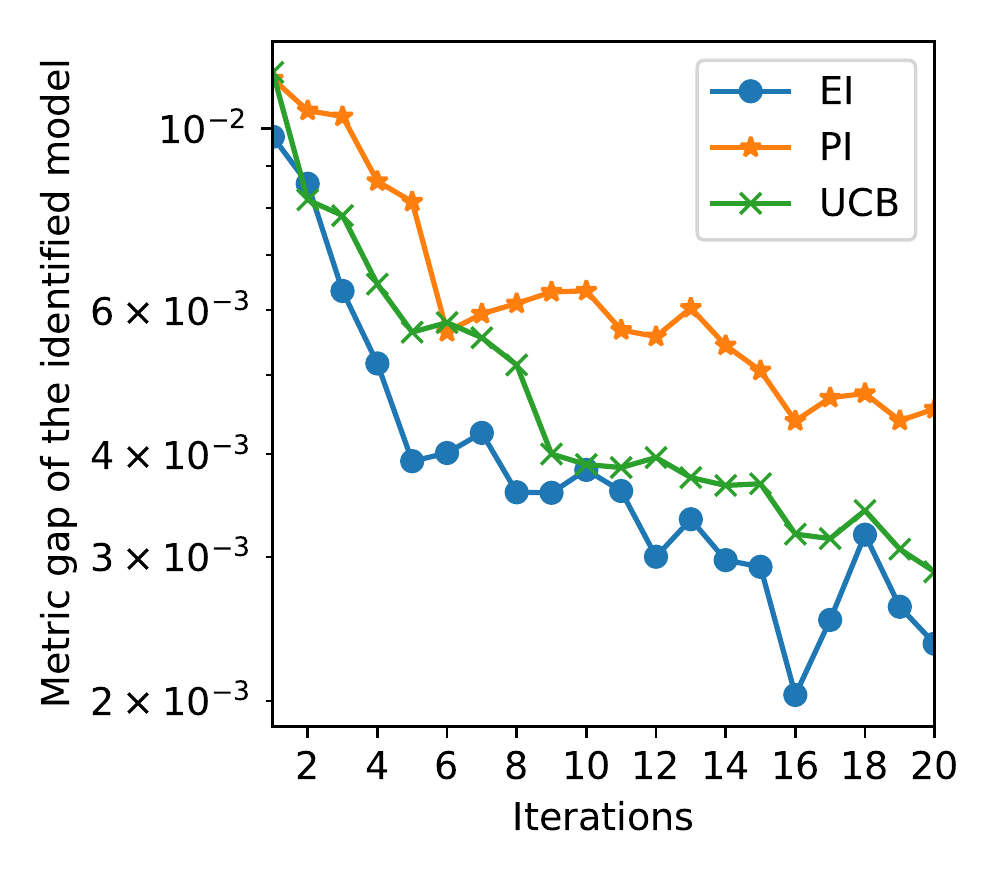}\vspace{-2.5mm}
         \caption{}
         \label{fig:svm_acq_progress}
     \end{subfigure}
     
       \begin{subfigure}[b]{\textwidth}
         \centering
         \includegraphics[width=\textwidth]{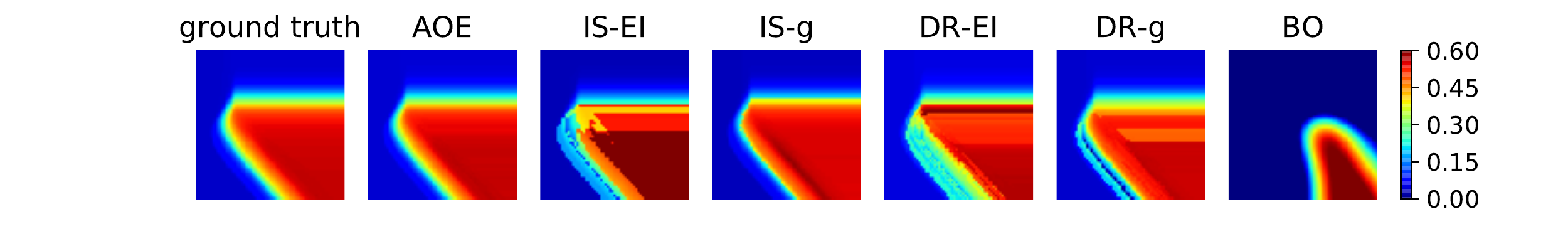}\vspace{-4mm}
         \caption{}
         \label{fig:svm_er_vis}
     \end{subfigure}
        \caption{Results of the classification experiment. (a) Comparison of \us{} and five baseline methods through the 20 sequential online experiments (refer to as iterations). The $y$-axis shows the gap in the accumulative metric between the optimal model and the estimated best model by each method. The average metric gaps after Iteration 20 are \textbf{0.0029}, 0.042, 0.024, 0.013, 0.020, 0.059 for \textbf{\us{}}, IS-EI, IS-g, DR-EI, DR-g, BO respectively. (b) RMSE of the estimated accumulative metrics for all the candidate models from each method. It is calculated from the same set of experiments as the ones in (a). The average RMSE after Iteration 20 are \textbf{0.011}, 0.061, 0.063, 0.044, 0.054, 1.31 for for \textbf{\us{}}, IS-EI, IS-g, DR-EI, DR-g, BO respectively.
        (c) Comparison of \us{} using different acquisition functions. (d) Heat maps of the estimated accumulative metric of all the candidate models after Iteration 20 comparing with the ground truth.}
\end{figure}

\textbf{Classification.} Inspired by how OPE works, we use a classification dataset to construct the first simulator. We consider the candidate models to be multi-class classifiers, and when deployed online, a model will be given a set of inputs but only receives binary feedback about whether the predicted class is correct. The task is to identify the model with the best accumulative metric in the smallest number of deployments, where the accumulative metric is the average accuracy on the hold-on set in this case. We use the ``letter" dataset from UCI repository~\citep{DuaGraff2017} and 
randomly take 200 data points for training and use the rest for ``online" experiments. 
In each online experiment, we randomly select 200 data points and pass them to the ``deployed" model and record the binary feedback and accumulative metric. 
We generate the set of candidate models by changing the two tuning parameter of support vector machine (SVM), $C$ and $\gamma$. The candidate model set is generated on a 100x100 grid in the space of the two parameters in log scale. In order to compare with the OPE-based baselines, all the decision $\aV$ is augmented with a $\epsilon$-greedy step with $\epsilon=0.05$.
We use a GP binary classifier with Mat\'{e}rn $\frac{3}{2}$ kernel as the surrogate model, using 2000 inducing points. We use EI as the acquisition function implemented in GPyOpt \citep{gpyopt2016}. Each run consists of 20 sequential online experiments with the first deployed model randomly picked. Each method repeatedly runs 20 times. 

Figure~\ref{fig:svm_progress} shows the comparison of all the methods as an average of 20 repeated runs. The performance at each iteration is measured as the gap in the accumulative metric between the optimal model and the estimated best model. 
The models picked by \us{} at all the iterations have significantly smaller metric gaps and the average metric gaps after Iteration 20 are \textbf{0.0029}, 0.042, 0.024, 0.013, 0.020, 0.059 for \textbf{\us{}}, IS-EI, IS-g, DR-EI, DR-g, BO respectively.
BO stops improving after about Iteration 10 because  it can only use the recorded accumulative metrics. DR-g and DR-EI are better than IS-g and IS-EI due to their lower variance estimator.
Figure~\ref{fig:svm_rmse_progress} shows the rooted mean square error (RMSE) between the estimated accumulative metrics for all the candidate models from each method, which is averaged across 20 runs. The average RMSE after Iteration 20 are \textbf{0.011}, 0.061, 0.063, 0.044, 0.054, 1.31 for for \textbf{\us{}}, IS-EI, IS-g, DR-EI, DR-g, BO respectively.
The RMSE of BO does not decrease due to the wrong generation from a few data points, which is worse than a flat prediction in the beginning. 
Figure~\ref{fig:svm_acq_progress} compares the different choices of acquisition functions (EI, PI, UCB) for \us{}, all of which performs similarly with EI being slightly better.
Figure~\ref{fig:svm_er_vis} shows the heat map visualization of the estimated accumulative metrics of all the candidate models after Iteration 20 from one of the 20 runs comparing with the ground truth. The $x$- and $y$-axis of the heat maps correspond to the two parameters $C$ and $\gamma$ in log scale. \us{} has the best visual resemblance to the ground truth among others, which is consistent with the RMSE result. 
See the supplement for the details.

\textbf{Recommender System.} We consider model selection for recommender system, which aims to select the best recommender  based on its online performance. 
In this experiment, a recommender system takes a user ID as input and returns an item ID for recommendation. For each recommendation, it receives binary feedback indicating whether the user has responded to the recommended item. The performance of a recommender system is measured by the average response rate, which is the accumulative metric.
We use the MovieLens 100k data \citep{HarperKonstan2015} to construct the simulator for online experiments.
Binary feedback is simulated according to the response probability that is computed by filling all the missing entries in the rating data with zero and mapping a 0-5 rating evenly into a probability between $[0.05, 0.95]$. 
We randomly take 20\% data for training and trained ten models using the Surprise package \citep{Nicolas2017}. In an online experiment, given a user ID, a trained model returns a predicted response probability and the recommendation is generated by taking the top five items and uniformly choosing one from them. The recommendation is augmented with $\epsilon$-greedy, where $\epsilon=0.05$.
Each run consists of five sequential online experiments. In each online experiment, every user is recommended five times.

\begin{figure}[t]
     \centering
     \begin{subfigure}[b]{0.4\textwidth}
         \centering
         \includegraphics[width=\textwidth]{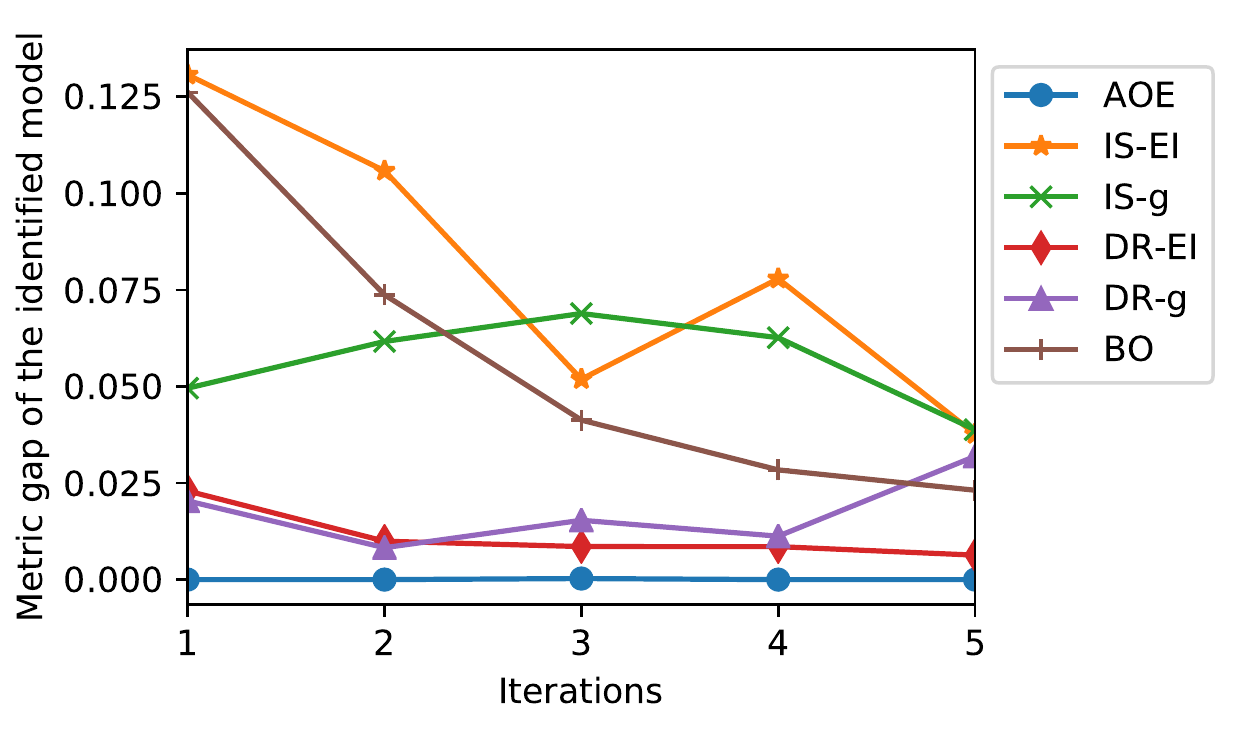}\vspace{-3mm}
         \caption{}
         \label{fig:movielens_progress}
     \end{subfigure}
~
     \begin{subfigure}[b]{0.4\textwidth}
         \centering
         \includegraphics[width=\textwidth]{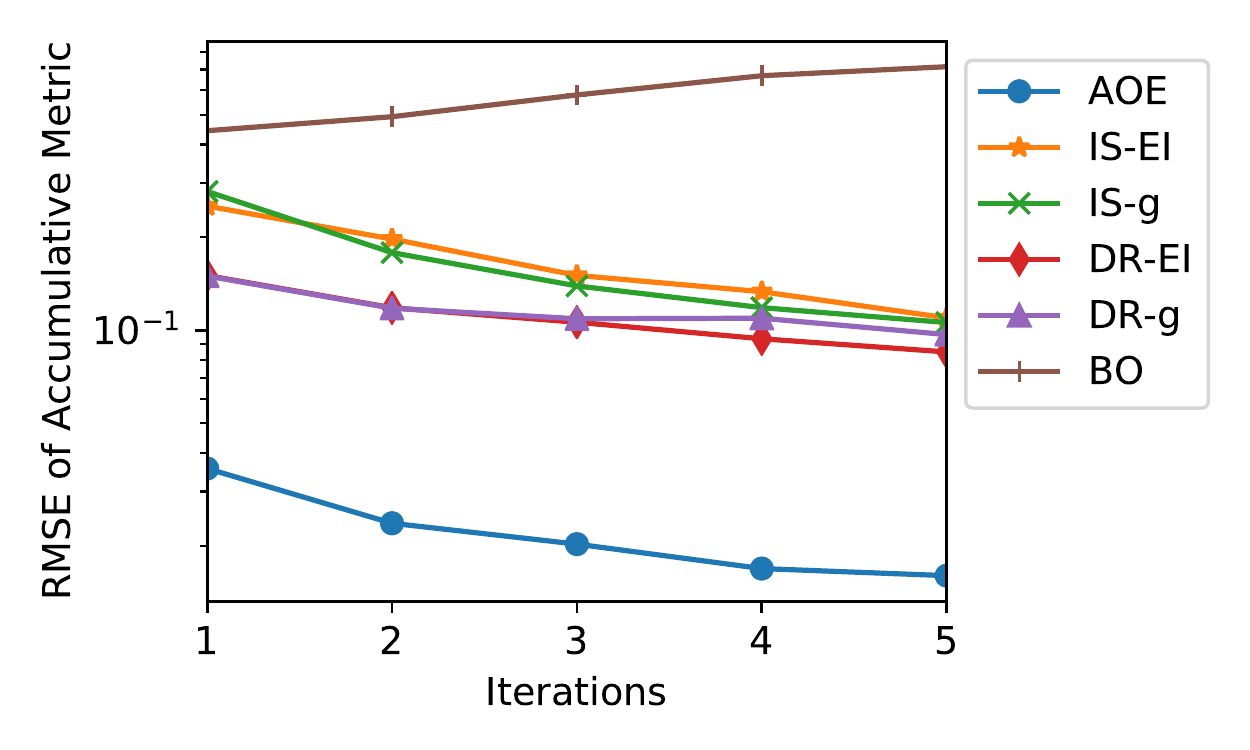}\vspace{-3mm}
         \caption{}
         \label{fig:movielens_rmse_progress}
     \end{subfigure}
        \caption{Results of the recommender system experiment. (a) Comparison of \us{} and five baseline methods through the five sequential online experiments. The $y$-axis shows the gap in the accumulative metric between the optimal model and the estimated best model by each method. The average metric gaps after Iteration 5 are \textbf{0.}, 0.038, 0.039, 0.0063, 0.032, 0.023 for \textbf{\us{}}, IS-EI, IS-g, DR-EI, DR-g, BO respectively. (b) RMSE of the estimated accumulative metrics of all the candidate models by each method. The average RMSE after Iteration 5 are \textbf{0.016}, 0.11, 0.11, 0.085, 0.097, 0.715 for for \textbf{\us{}}, IS-EI, IS-g, DR-EI, DR-g, BO respectively.
        }
        \label{fig:recsys}
\end{figure}

Figure~\ref{fig:recsys} shows the comparison with the average of 20 repeated runs. As shown in Figure~\ref{fig:movielens_progress}, \us{} always identifies the best model from the first iteration and BO fails to reduce the RMSE because the surrogate model of BO is not aware of the objective ranged between zero and one. 
The average metric gaps after Iteration 5 are \textbf{0.}, 0.038, 0.039, 0.0063, 0.032, 0.023 for \textbf{\us{}}, IS-EI, IS-g, DR-EI, DR-g, BO respectively.
In Figure~\ref{fig:movielens_rmse_progress}, we show the RMSE of the estimated accumulative metrics. 
The RMSE from \us{} continuously decreases, while it always correctly identifies the best model.  
The average RMSE after Iteration 5 are \textbf{0.016}, 0.11, 0.11, 0.085, 0.097, 0.715 for for \textbf{\us{}}, IS-EI, IS-g, DR-EI, DR-g, BO respectively.

\section{Discussion}

The model selection for production system does not fit into the classical model selection paradigm.  
We propose a new approach by taking data collection into the model selection process and selecting the best model via iterative online experiments. It allows selection from a much larger pool of candidates than using A/B testing and gives more accurate selection than off-policy evaluation by actively reducing selection bias.
We design a GP surrogate model for predicting immediate feedback and derive the distribution of the accumulative metric.
The model to deploy at each iteration is picked by balancing the predicted accumulative metric and the uncertainty of the prediction due to limited data. 
With simulated experiments from real data, we show that \us{} performs significantly better than all the baselines in terms of identifying the best model and estimating the accumulative metric.

The concept of iterative model deployment also appear in bandit algorithms. A bandit algorithm performs exploration-exploitation for individual user interactions and continuously updates the model. A major difference to our paradigm is that a bandit algorithm is often applied to the decision making scenarios that have tight time constraints because of its short decision time, in which the model is either not updated or updated with incremental learning after each action, while our method selects among different candidate models and each online experiment contains lots of user interactions, which allows us to consider more expensive surrogate models and retrain the surrogate model after each action.
RL as a broader framework also considers the problem about evaluating and updating a policy from recorded data with respect to a generic form of reward, which may be delayed and depend on sequential actions. \problem{} can be viewed as a special case of RL, in which the reward is an average of immediate feedback. This special setting allows us to develop a dedicated surrogate model and make efficient use of data, which is not applicable to the generic RL setting. On the other hand, RL offers an interesting future direction for handling broader types of accumulative metric beyond the form of average.

\section*{Broader Impact}

In this paper, the authors present a new framework of model selection for production system (\problem{}), in which the data collection from automated online experiments is used as part of the model selection procedure. In particular, we develop \us{}, a \problem{} method that iteratively select models to be deployed online and identify the model with the highest metric of interest from a large pool of candidate models in a few number of online deployments.

\us{} could be applied to improve the quality of the model selection process for industrial ML service development. This type of methods could be implemented  either as a part of the in-house development toolset of individual companies or as a component of automated ML service on cloud platforms. The adoption of such tooling could increase the development speed of industrial ML applications and provide better understanding and control of the release of new features and improvement before large scale deployment. With a more accurate prediction of the online metric of a system improvement, ML developers can better identify the impactful system improvements and focus the development effort on them. It also can let the development team and a wider part of a company have a clear picture of the potential impact and limitation of a project before the development has finished.

The automated ML model selection, update, deployment tools including \us{} tend to focus on a single metric for mathematical convenience, but the social impact of a ML system such as diversity, fairness is hard to summarize into a single metric. The adoption of such tooling without careful consideration can result into overly optimize for the single metric and being blind about broad social impacts, which can potentially lead to undesirable outcomes. The research about understanding and constraining automated algorithm decisions with respect to its wider impact, \eg{}, safe reinforcement learning, is very important and could mitigate the risk of causing harmful consequences.

\bibliography{bops}

\newpage
\appendix
\section*{\Large \centering Supplemental Material for ``Model Selection for Production System via Automated Online Experiments''}
\vspace{3mm}

\section{Experiment Details}

In the following section, we will present the additional details about our experiments that do not fit in the main text.

\subsection{Baseline methods}

As \problem{} is a new framework for model selection, we construct five baseline methods by extending the related method into our scenario and compare with \us{}. The five baseline methods are as follows:

\begin{description}
\item [BO] For each online experiment, we can have an unbiased estimate of the accumulative metric under the deployed model as mentioned in Section~\ref{sec:problem_statement}. We directly apply Bayesian Optimization (BO) to the model selection problem by taking the set of candidate models as the input space and the estimate of the accumulative metric from online experiments as the output and treating \problem{} as an optimization problem. We use the default setting of BO in GPyOpt, where the surrogate model is a Gaussian process (GP) regression model with a Gaussian noise distribution and a M\'atern 5/2 kernel. Expected Improvement (EI) is used as the acquisition function. For the classification experiment, as the candidate models are naturally generated from a 2D space of the SVM parameters C and $\gamma$, we use the values of these two parameters to identify individual candidate models and use this 2D space as the search space for BO. However, for the recommender system experiment, there are no natural representations for the candidate models. We treat each candidate model as a categorical value, which leads to its bad performance.
\item [IS-g / DR-g] Off-policy evaluation (OPE) methods can provide an estimate of the accumulative metric. We use two popular OPE methods, importance sampling (IS) and doubly robust (DR) to estimate the accumulative metric after each online experiment and greedily choose the candidate model with the highest estimated accumulative metric for the next online experiment. We denote the resulting two methods as IS-g and DR-g respectively. 
\item [IS-EI / DR-EI] IS-g and DR-g suffer from the fact that there is no exploration mechanism. To offer a stronger baseline, we not only use IS and DR to estimate the accumulative metric, but also calculate the empirical variance of the resulting estimate. Then, we score the candidate models according to an acquisition function (EI is used in the experiments) and select the next model to deploy with the highest score. The resulting methods are denoted as IS-EI and DR-EI respectively.
\end{description}

As there are limited information to be gained by repeatedly deploying the same model online, we exclude the models that have been deployed when choosing the next model to deploy for all the methods including \us{}.

\subsection{Classification}

We take the inspiration from the OPE literature and construct an online experiment scenario using a classification dataset. We simulate the ``online'' deployment scenario as follows: a multi-class classifier is given a set of inputs; for each input, the classifier returns a prediction of the label and only a binary immediate feedback about whether the predicted class is correct is available. The performance of a classifier is measured by the average accuracy on the hold-out set, which corresponds to the accumulative metric. As only one model can be deployed at a time and in each deployment a small subset of the hold-out data are used, the task is to select the best-performing model in the smallest number of deployments. 

We use the ``letter" dataset from UCI repository~\citep{DuaGraff2017}. There are, in total, 20,000 data points in the dataset.  
We randomly sample 200 data points for training and use the rest for ``online" experiments. In each online experiment, we randomly select 200 data points from the hold-out set and pass them to the ``deployed" model and record the binary feedback and accumulative metric. 

We consider support vector machine (SVM) as the multi-class classifier and generate the set of candidate models by varying the two tuning parameters of SVM for training, $C$ and $\gamma$. 
To demonstrate that \us{} can select a good model from a large set of candidates, we generate in total 10,000 candidate models by choosing $C$, and $\gamma$ from a 100x100 grid in the space of these two parameters in log scale. We follow the guideline from previous works and consider $C$ between $2^{-5}$ and $2^{15}$ and $\gamma$ between $2^{-15}$ and $2^3$.
In order to compare with the OPE-based baselines, all the predictions from an SVM are augmented with a $\epsilon$-greedy step, \ie{}, the predicted label is sampled according to a categorical distribution, in which the class predicted by the SVM has $1-\epsilon$ probability and the rest classes evenly share the probability $\epsilon$. We set $\epsilon=0.05$.

We use a GP binary classifier with a Mat\'{e}rn $\frac{3}{2}$ kernel as the surrogate model. We use 2000 inducing points and EI as the acquisition function. When training the surrogate model, we use Adam as the gradient optimizer for variational inference, which runs for 600 epochs with the mini-batch size being 100 and the learning rate being 0.001 with stratified sampling. We use logistic regression as the baseline model for DR-based baselines.

Each model selection experiment consists of 20 sequential online experiments and the model deployed in the first experiments is randomly picked according to a uniform distribution for all the methods.  We repeatedly run 20 experiments for each model. In each repeated run, the set of candidate models are the same but the first deployed model and the data points sampled for each online experiment may be different due to random sampling.

\begin{figure}[t]
     \centering
     \begin{subfigure}[b]{0.49\textwidth}
         \centering
         \includegraphics[width=\textwidth]{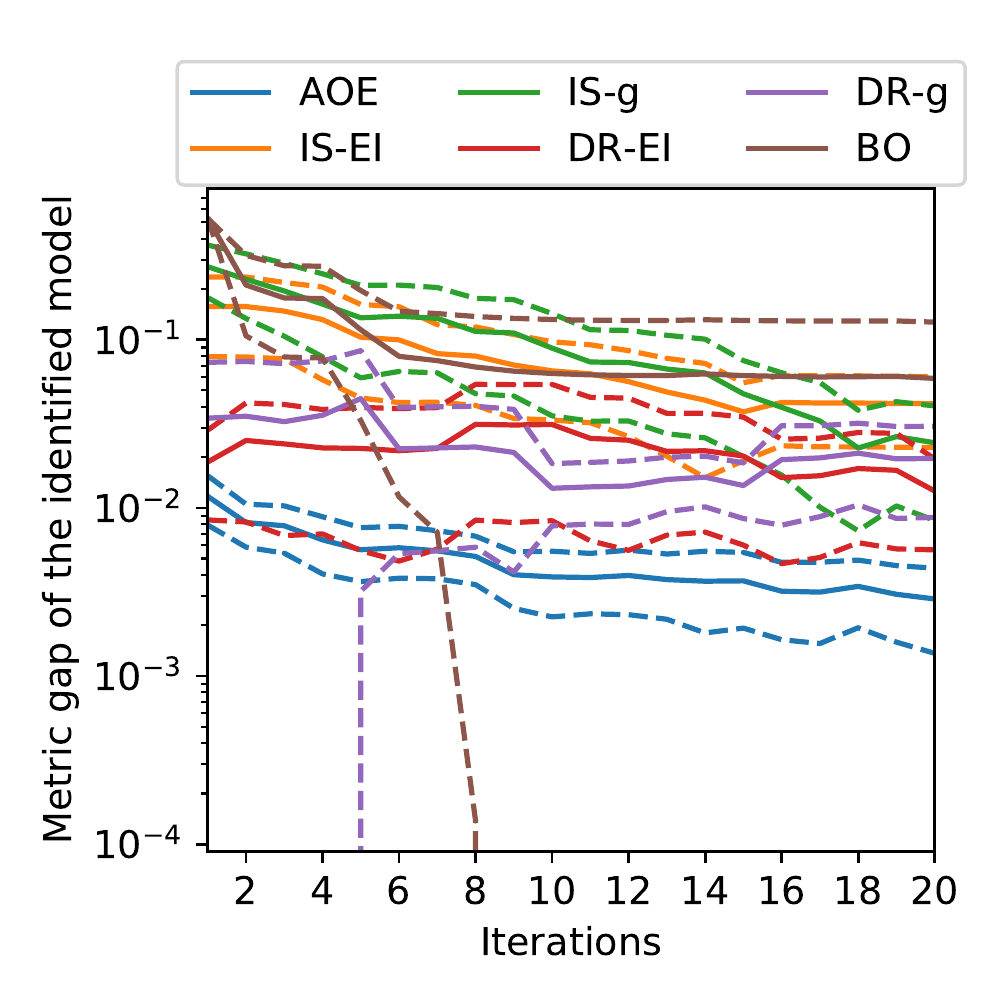}
         \caption{}
         \label{fig:svm_progress_errorbar}
     \end{subfigure}
     ~
     \begin{subfigure}[b]{0.49\textwidth}
         \centering
         \includegraphics[width=\textwidth]{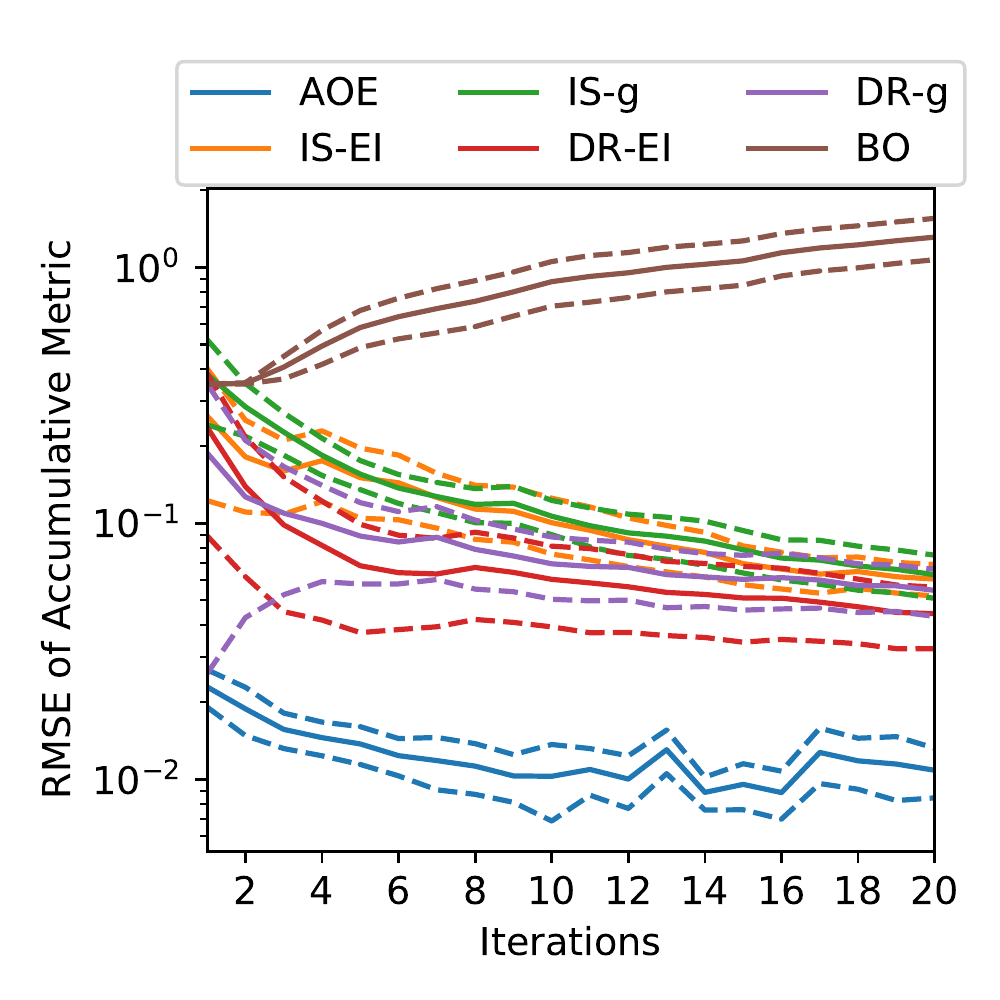}
         \caption{}
         \label{fig:svm_rmse_progress_errorbar}
     \end{subfigure}
        \caption{Additional results of the classification experiment. (a) Comparison of \us{} and five baseline methods through the 20 sequential online experiments (refer to as iterations). The $y$-axis shows the gap in the accumulative metric between the optimal model and the estimated best model by each method.  (b) RMSE of the estimated accumulative metrics for all the candidate models from each method. The error bars in both (a) and (b) indicate the confidence interval of the estimated mean by two times of the standard deviation.}
\end{figure}

\begin{figure}[t]
     \centering
     \begin{subfigure}[b]{0.49\textwidth}
         \centering
         \includegraphics[width=\textwidth]{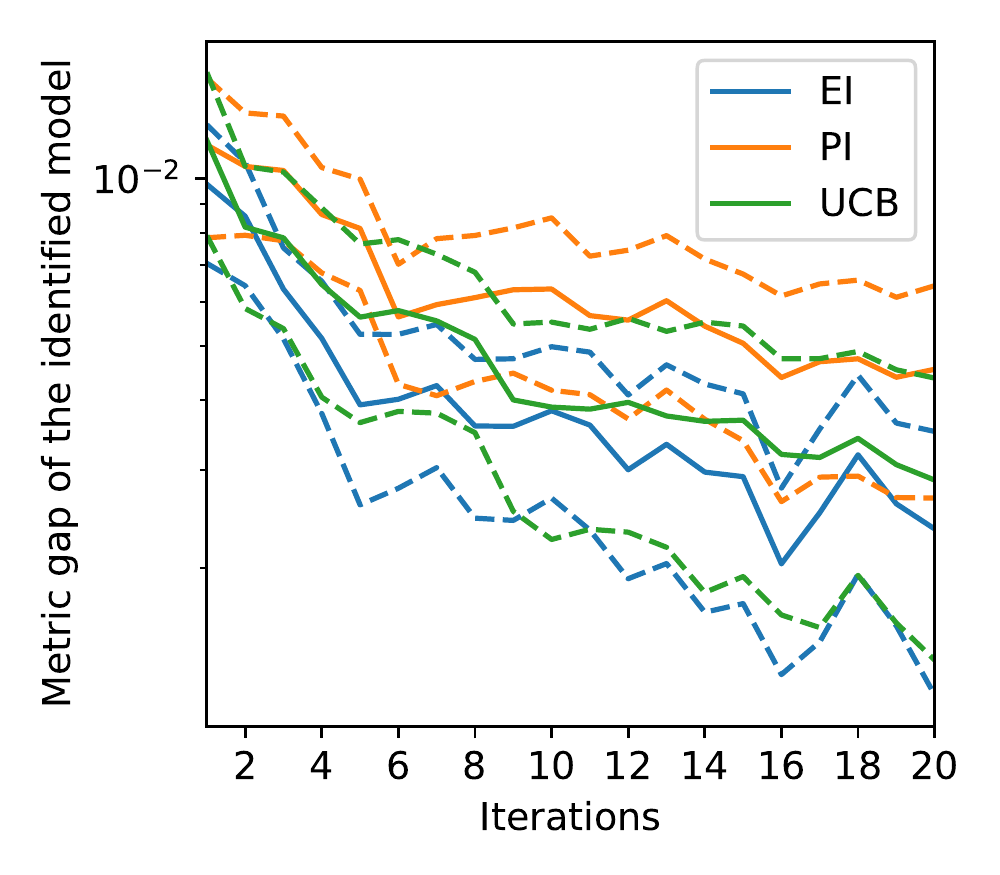}
         \caption{}
         \label{fig:svm_acq_progress_errorbar}
     \end{subfigure}
     ~
     \begin{subfigure}[b]{0.49\textwidth}
         \centering
         \includegraphics[width=\textwidth]{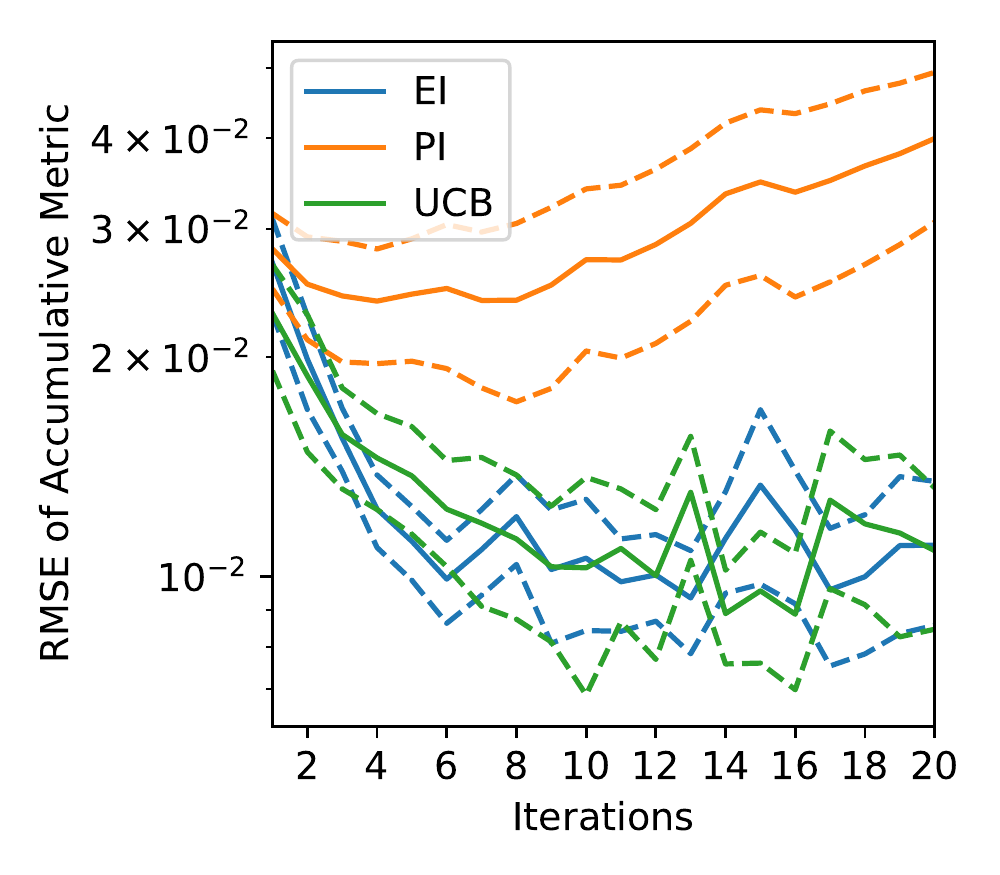}
         \caption{}
         \label{fig:svm_acq_rmse_progress_errorbar}
     \end{subfigure}
        \caption{Additional results about the comparison of acquisition functions. It compares the performance of different acquisition function used by \us{} in the classification experiment. (a) Comparison of different acquisition functions in terms of the gap in the accumulative metric between the optimal model and the estimated best model. (b) Comparison of different acquisition function in terms of RMSE of the estimated accumulative metrics. The error bars in both (a) and (b) indicate the confidence interval of the estimated mean by two times of the standard deviation.}
\end{figure}

\begin{table}
\centering
\caption{The metric gap and RMSE after Iteration 20 in the classification experiment}
\label{tab:classification}
\begin{tabular}{ |c|c|c| } 
 \hline
 & Metric Gap & RMSE \\ \hline\hline
 \us{} & \textbf{0.0029 (0.0033)} & \textbf{0.011 (0.0053) } \\ \hline
IS-EI & 0.042 (0.042) & 0.061 (0.019) \\ \hline
IS-g & 0.024 (0.036) & 0.063 (0.027) \\ \hline
DR-EI & 0.013 (0.016) & 0.044 (0.027) \\ \hline
DR-g & 0.020 (0.024) & 0.054 (0.026) \\ \hline
BO & 0.059 (0.15) & 1.31 (0.54) \\ 
 \hline
\end{tabular}
\end{table}

Figure~\ref{fig:svm_progress_errorbar} shows the comparison of all the methods with error bars in terms of the the gap in the accumulative metric between the optimal model and the estimated best model. 
Figure~\ref{fig:svm_rmse_progress_errorbar} shows the average rooted mean square error (RMSE) of the estimated accumulative metrics after each iteration with error bars.
The error bars indicate the confidence interval of the estimated mean by two times of the standard deviation.
The average metric gaps and average RMSE of all the methods after Iteration 20 are shown in Table~\ref{tab:classification}. The values in the parentheses indicates the standard deviation of the metric gap and RMSE across the 20 repeated runs.

Apart from the comparison between \us{} and the baseline methods. We also compare the performance of \us{} when using different acquisition functions. Figure~\ref{fig:svm_acq_progress_errorbar} shows the comparison of different acquisition functions with error bars in terms of the the gap in the accumulative metric between the optimal model and the estimated best model. Figure~\ref{fig:svm_acq_rmse_progress_errorbar} shows the average RMSE of the estimated accumulative metrics after each iteration with error bars with different acquisition functions. The error bars indicate the confidence interval of the estimated mean by two times of the standard deviation.

\begin{sidewaysfigure}[ht]
    \includegraphics[width=\textwidth]{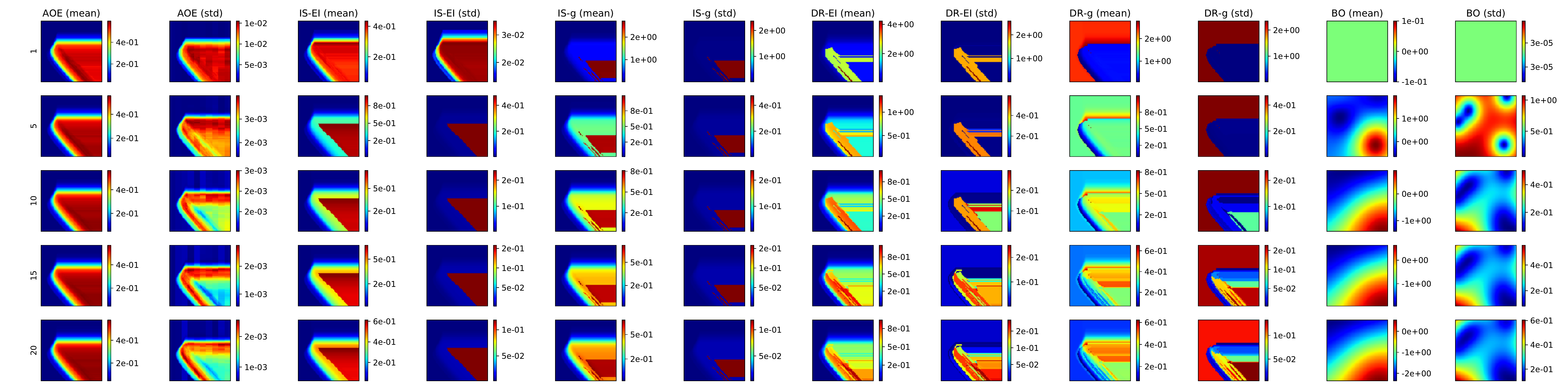}
    \caption{The heat map visualization of the estimated accumulative metrics of all the candidate models after Iteration 1, 5, 10, 15 and 20. The results come from one of the 20 repeated runs. The $x$- and $y$-axis of the heat maps correspond to the two parameters $C$ and $\gamma$ in log scale. Each column corresponds to a method. ``(mean)'' indicates the visualization of the mean of the estimation and ``(std)'' indicates the visualization of the standard deviation estimated by each method.}
    \label{fig:svm_er_vis_full}
\end{sidewaysfigure}

To illustrate the behaviors of \us{} and the baseline methods during the model selection process, we visualize the mean and standard deviation of the estimated accumulative metrics after Iteration 1, 5, 10, 15, and 20 from individual methods in Figure~~\ref{fig:svm_er_vis_full}. The visualization uses one of the 20 runs. Note that, to provide more information, Figure~\ref{fig:svm_er_vis}  and Figure~\ref{fig:svm_er_vis_full} use different runs.

\subsection{Recommender System}

\begin{figure}[t]
     \centering
     \begin{subfigure}[b]{0.49\textwidth}
         \centering
         \includegraphics[width=\textwidth]{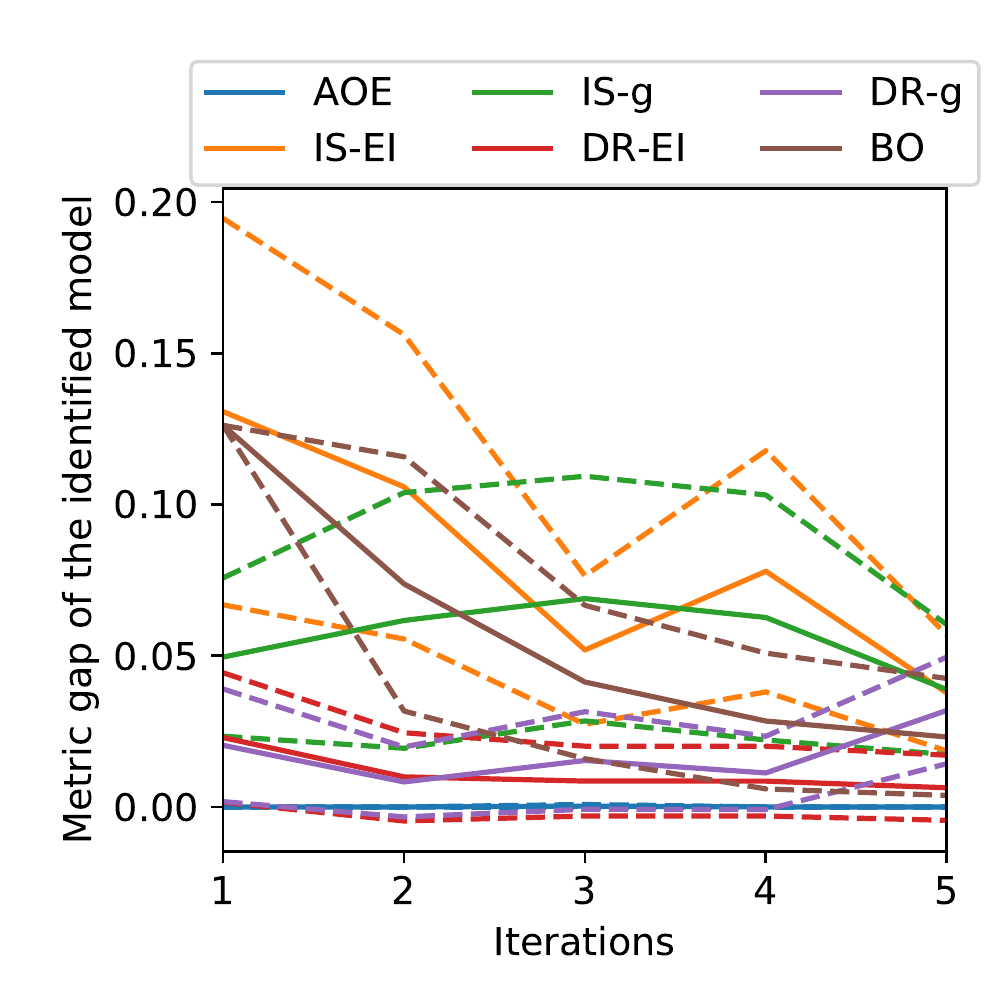}
         \caption{}
         \label{fig:movielens_progress_errorbar}
     \end{subfigure}
     ~
     \begin{subfigure}[b]{0.49\textwidth}
         \centering
         \includegraphics[width=\textwidth]{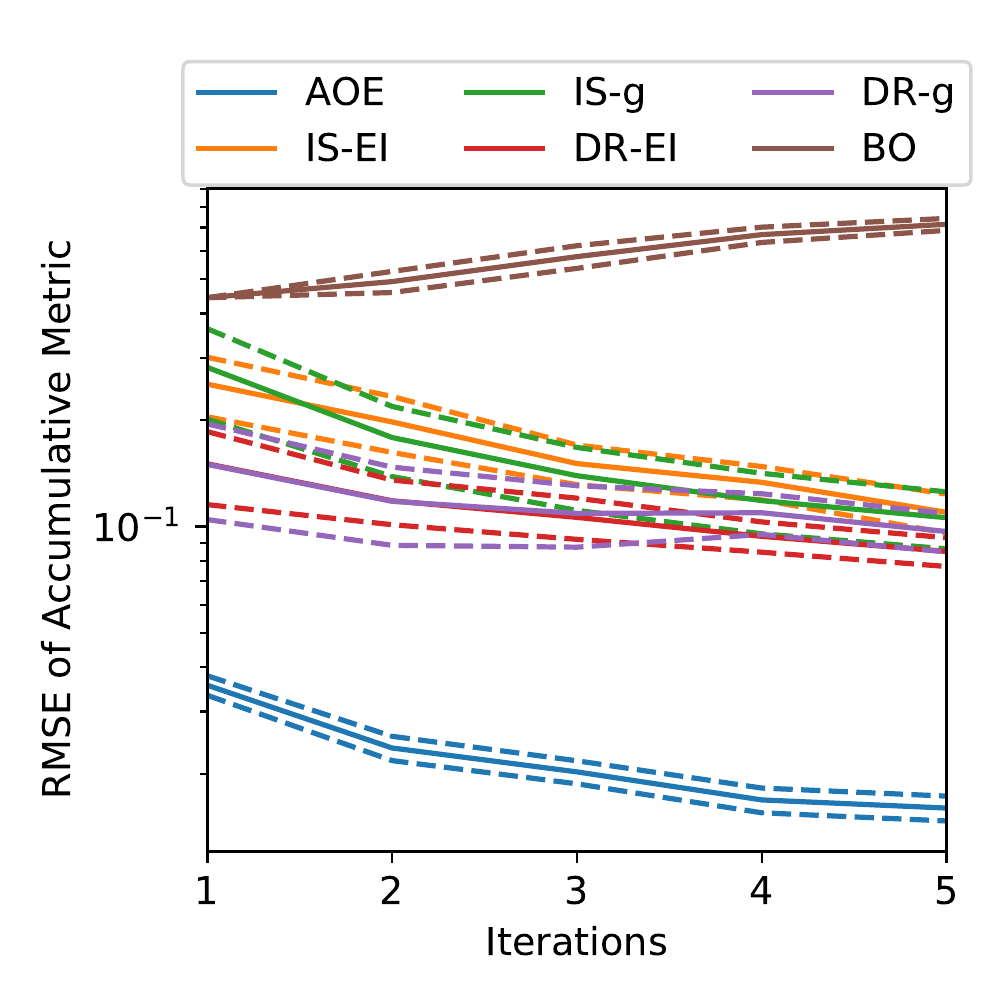}
         \caption{}
         \label{fig:movielens_rmse_progress_errorbar}
     \end{subfigure}
        \caption{Additional results of the recommender system experiment. (a) Comparison of \us{} and five baseline methods through the 20 sequential online experiments (refer to as iterations). The $y$-axis shows the gap in the accumulative metric between the optimal model and the estimated best model by each method.  (b) RMSE of the estimated accumulative metrics for all the candidate models from each method. The error bars in both (a) and (b) indicate the confidence interval of the estimated mean by two times of the standard deviation.}
\end{figure}

\begin{table}
\centering
\caption{The metric gap and RMSE after Iteration 5 in the recommender system experiment}
\label{tab:recsys}
\begin{tabular}{ |c|c|c| } 
 \hline
 & Metric Gap & RMSE \\ \hline\hline
 \us{} & \textbf{0. (0.)} & \textbf{0.016 (0.0029) } \\ \hline
IS-EI & 0.038 (0.043) & 0.11 (0.031) \\ \hline
IS-g & 0.039 (0.048) & 0.11 (0.043) \\ \hline
DR-EI & 0.0063 (0.024) & 0.085 (0.018) \\ \hline
DR-g & 0.032 (0.039) & 0.097 (0.027) \\ \hline
BO & 0.023 (0.043) & 0.715 (0.062) \\ 
 \hline
\end{tabular}
\end{table}

We demonstrate \us{} on the problem of model selection for recommender system, which aims to select the best recommender based on its online performance. 
In this experiment, we consider that a recommender system takes a user ID as input and returns an item ID for recommendation. 
For each recommendation, it receives binary feedback indicating whether the user has responded to the recommended item. 
We measure the performance of such a recommender system by the average response rate, which corresponds to the accumulative metric.
We construct a simulator by using the MovieLens 100k data \citep{HarperKonstan2015}. Given a user ID and an item ID, the binary feedback is simulated by drawing a sample from a Bernoulli distribution, in which the probably of being one is specified by the response probability corresponding to the user ID and item ID pair. 
The MovieLens 100k data provide the ratings corresponding to a list of user and item pairs. 
We filter items that have average rating below a threshold of 0.2.
The ratings range between one and five. We generate a full table of the response probability for all the user and item combinations by first filling all the missing entries in the rating data with zero and mapping the resulting 0-5 rating evenly to a probably between $[0.05, 0.95]$, \ie{}, 0.05 for 0, 0.23 for 1, 0.41 for 2, 0.59 for 3, 0.77 for 4 and 0.95 for 5.

We do not use any user and item features and randomly take 20\% of the entries in the response probability table for training the candidate models. We trained ten models using the Surprise package \citep{Nicolas2017} with their default setting. The full list of the names of the models are SVD, BaselineOnly, CoClustering, KNNBaseline, KNNWithMeans, NormalPredictor, NMF, KNNWithZScore, KNNBasic, SlopeOne. At prediction time, each of these models predicts a response probability given a pair of user ID and item ID.
In an online experiment, given a user ID, a trained model predicts the response probabilities of all the items, and the recommendation is generated by taking the top five items and randomly choosing one from them following a uniform distribution. The recommendation is augmented with $\epsilon$-greedy, \ie{}, the item for recommendation is sampled from a categorical distribution, in which the top five items have $(1-\epsilon)/5$ probability and the rest items evenly share $\epsilon$ probability. We set $\epsilon=0.05$.

The users and items are represented by their IDs, which are not good representations for GP. We augment a GP binary classifier by embedding the user and item IDs into two separate latent spaces as mentioned in Sec.~\ref{sec:gp_surrogate_model} and use it as the surrogate model. We use 5D latent spaces for the user and item embedding separately and an RBF kernel. We use 1000 inducing points and EI as the acquisition function. When training the surrogate model, we use Adam as the gradient optimizer for variational inference, which runs for 200 epochs with the mini-batch size being 100 and the learning rate being 0.001. As the prediction task can also be viewed as matrix imputation, we use the KNNImputer from the scikit-learn package as the predictive model for DR-based baselines.

Each model selection experiment consists of five sequential online experiments and the model deployed in the first experiments is randomly picked according to a uniform distribution for all the methods. The data collected in each online experiment are generated by considering each user for recommendation five times. We repeatedly run 20 experiments for each model. In each repeated run, the set of candidate models is the same but the first deployed model and the data points sampled for each online experiment may be different.

\begin{figure}[t]
     \centering
         \includegraphics[width=\textwidth]{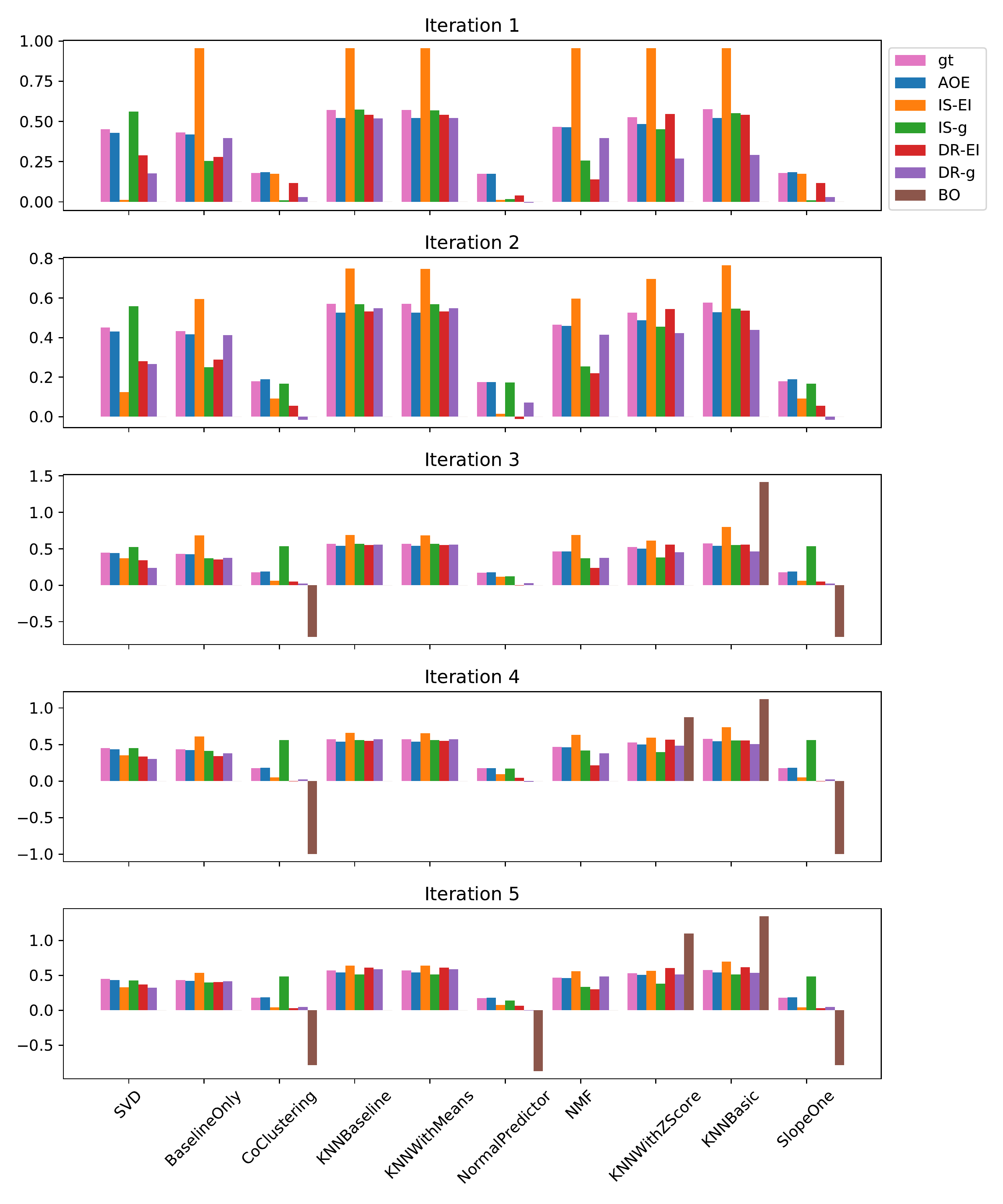}
         \caption{The bar plot of the estimated accumulative metrics of all the candidate models after each iteration comparing with ground truth (denoted as ``gt''). The results come from one of the 20 repeated runs. The $y$-axis shows the accumulative metric. In the $x$-axis, each group of bars corresponds to a candidate model (there are ten in total.) and each color of bars corresponds to all the compared methods plus the ground truth.}
         \label{fig:movielens_er_vis_full}
\end{figure}

Figure~\ref{fig:movielens_progress_errorbar} shows the comparison of all the methods with error bars in terms of the gap in the accumulative metric between the optimal model and the estimated best model. 
Figure~\ref{fig:movielens_rmse_progress_errorbar} shows the average RMSE of the estimated accumulative metrics after each iteration with error bars.
The error bars indicate the confidence interval of the estimated mean by two times the standard deviation.
The metric gaps and RMSE of all the methods after Iteration 5 are shown in Table~\ref{tab:recsys}.
To illustrate the behaviors of \us{} and the baseline methods during the model selection process, we visualize the estimated accumulative metric after each iteration comparing with the ground truth in Figure~~\ref{fig:movielens_er_vis_full}. The visualization uses one of the 20 runs.

\section{Details about Sparse GP and Variational Inference}

For scalability, we use the variational sparse GP approximation to speed up the inference. Variational sparse GP augments the original data with a set of pseudo data $\uV$ at the corresponding locations $\zM$, which shares the same GP as in the original model. The input locations of the pseudo data $\zM$ lie in the joint space of the action and input $\aV$ and $\xV$. The resulting model is a joint GP between the original data and the augmented data $p(\fV, \uV| \aM, \xM, \zM)$, which can also be written as a product of the conditional distribution $p(\fV | \uV, \aM, \xM, \zM)p(\uV | \zM)$. Note that such an augmentation does not change the original model,
\begin{equation}
p(\fV|\aM, \xM) = \int p(\fV|\uV, \aM, \xM, \zM) p(\uV|\zM) \dif \uV.
\end{equation}
Both $\fV$ and $\uV$ are latent variables. Variational sparse GP assumes a specific variational posterior distribution $q(\fV, \uV) = p(\fV|\uV) q(\uV)$, where $q(\uV) = \mathcal{N}(\mV_{\uV}, \sM_{\uV})$ is a multi-variate normal distribution, of which the mean and covariance matrix are variational parameters. A variational lower bound can be derived with the above variational posterior, of which the computational complexity reduces from $O(N^3)$ to $O(NC^2)$, where $C$ is the number of pseudo data. More details about the variational approximation can be found in \citep{Titsias2009}.

After inferring the variational posterior of the sparse GP, the distribution of the accumulative metric conditioned on the observed data can be derived based on the variational posterior of sparse GP,
\begin{equation}
p(\hat{v}|\model_i, \mathcal{D}) = \mathcal{N}\left( \frac{1}{T}\pM_: ^\top  \K_{*\uV} \K_{\uV\uV}^{-1} \mV_{\uV}, \frac{1}{T} \pM_:^\top ( \K_{**} - \K_{*\uV} (\K_{\uV\uV}^{-1} - \K_{\uV\uV}^{-1}  \sM_{\uV} \K_{\uV\uV}^{-1} )\K_{*\uV}^\top)\pM_: \right),
\end{equation}
where $\K_{\uV\uV}$ is the covariance matrix among the pseudo data, $\K_{*\uV}$ is the cross-covariance matrix between $\wM$ and the pseudo data, and $\mV_{\uV}$ and $\sM_{\uV}$ are the inferred variational parameters in $q(\uV)$. With this above derived distribution of the accumulative metric, we can apply an acquisition function for selecting a candidate model.

For a large problem, the variance calculation in the above distribution can also be very expensive as  $\K_{**}$ is a $KT$-by-$KT$ matrix. For efficient computation, we apply a FITC approximation \citep{SnelsonGhahramani2006} at prediction time, 
\begin{equation}
p_{\text{FITC}}(\fV|\uV, \aM, \xM, \zM) = \mathcal{N}(\K_{\fV\uV}\K_{\uV\uV}^{-1}\uV, \mathbf{\Lambda} ),
\end{equation}
where $\mathbf{\Lambda} =\diag{\K_{\fV\fV} -\K_{\fV\uV}\K_{\uV\uV}^{-1} \K_{\fV\uV}^\top}$ and  $\diag{\cdot}$ makes a matrix into a diagonal matrix by letting off-diagonal entries be zero. Note that, although the conditional distribution $p(\fV|\uV)$ is independent among the entries of $\fV$, the resulting distribution $p(\mVbar | \aM, \xM, \mathcal{D})$ is still correlated due to the correlation from the pseudo data.
With the FITC approximation, the resulting distribution of the accumulative metric becomes
\begin{equation}
p(\hat{v}|\model_i, \mathcal{D}) = \mathcal{N}\left( \frac{1}{T}\pM_: ^\top  \K_{*\uV} \K_{\uV\uV}^{-1} \mV_{\uV}, \frac{1}{T} \pM_:^\top ( \mathbf{\Lambda} + \K_{*\uV}  \K_{\uV\uV}^{-1}  \sM_{\uV} \K_{\uV\uV}^{-1} \K_{*\uV}^\top)\pM_: \right),
\end{equation} 
where only the diagonal entries of $\K_{**}$  needs to be computed.

\end{document}